\documentclass[a4paper]{article}

\usepackage{INTERSPEECH2019}
\usepackage{url}

\title{The Zero Resource Speech Challenge 2019: TTS without T}

\name{Ewan Dunbar$^{1,2*}$,
      Robin Algayres$^2$,
      Julien Karadayi$^2$,
      Mathieu Bernard$^{2}$,
      Juan Benjumea$^2$,
      Xuan-Nga Cao$^{2}$,
      Lucie Miskic$^1$,
      Charlotte Dugrain$^1$,
      Lucas Ondel$^3$,\\
      Alan W. Black$^4$,
      Laurent Besacier$^5$,
      Sakriani Sakti$^{6,7}$,
      Emmanuel Dupoux$^{2,8}$}
\address{
  $^1$Laboratoire de Linguistique Formelle (CNRS - Paris Diderot - Sorbonne Paris Cit\'e), France\\
  $^2$Cognitive Machine Learning (ENS - CNRS - EHESS - INRIA - PSL Research University), France\\
  $^3$Department of Computer Graphics and Multimedia, Brno Univ. of Technology, Czech Republic\\
  $^4$Language Technologies Institute, Carnegie Mellon University, USA\\
  $^5$Laboratoire d'Informatique de Grenoble, \'equipe GETALP (Universit\'e Grenoble Alpes), France\\
  $^6$Nara Institute of Science and Technology, $^7$RIKEN Center for Advanced Intelligence Project, Japan\\
  $^8$Facebook A.I. Research, Paris, France
}
\email{$^*$ewan.dunbar@univ-paris-diderot.fr}

\begin{document}

\maketitle


\begin{abstract}
We present the Zero Resource Speech Challenge 2019, which 
proposes to build a speech synthesizer without any text or phonetic labels: hence, TTS without T (text-to-speech without text). We provide raw audio for a target voice in an unknown language (the Voice dataset), but no alignment, text or labels. Participants must discover subword units in an unsupervised way (using the Unit Discovery dataset) and align them to the voice recordings in a way that works best for the purpose of synthesizing novel utterances from novel speakers, similar to the target speaker's voice. We describe the metrics used for evaluation, a baseline system consisting of unsupervised subword unit discovery plus a standard TTS system, and a topline TTS using gold phoneme transcriptions. We present an overview of the 19 submitted systems from 10 teams and discuss the main results.

\end{abstract}

\section{Introduction}

Young children learn to talk long before they learn to read and write. They can produce novel sentences without being trained on  speech annotated with text. Presumably, they achieve this by encoding input speech in their internal phonetic speaker-invariant representations (proto-phonemes), and use this representation to generate speech in their own voice. Reproducing this ability would be useful for the thousands of so-called low-resource languages, which lack the textual resources and/or expertise required to build traditional  synthesis systems. 

\begin{figure}[t]
  \centering
  \includegraphics[width=.8\linewidth]{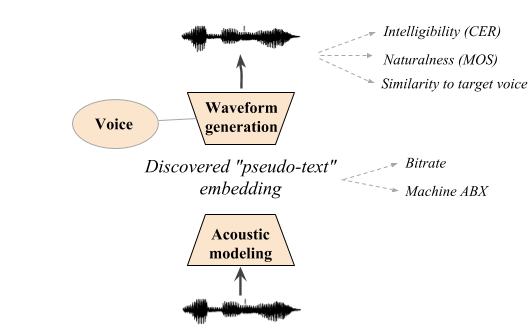}
  \caption{Schematic diagram of the challenge.}
  \label{fig:diagram-challenge}
\end{figure}

The Zero Resource Speech Challenge 2019 (ZR19: \mbox{\url{www.zerospeech.com/2019/}}) proposes to build a speech synthesizer without text or labels. We provide raw audio for the target voice(s) in an unknown language, but no text or labels. Participants must discover subword units in an unsupervised way and align incoming speech to these units in a way that allows for synthesizing novel utterances from novel speakers (see Figure \ref{fig:diagram-challenge}). It is a continuation of the sub-word unit discovery track of ZeroSpeech 2017 \cite{versteegh2016zero} and ZeroSpeech 2015 \cite{dunbar2017zero}, as it demands of participants to discover such units, and then evaluate them by assessing their performance on a novel speech synthesis task.

 As with the other two challenges, it relies exclusively on freely accessible software and datasets. We provide a baseline system which performs the task using two off-the-shelf components: (a) a system which discovers discrete acoustic units automatically, and (b) a standard TTS system. A submission to the challenge replaces at least one of these systems. The challenge is  therefore open to systems which make a contribution primarily to unit discovery, as well as to TTS-only systems which concentrate primarily on improving the quality of the synthesis on the baseline sub-word units. Participants can of course also construct their own end-to-end system with the  objective of discovering sub-word units and producing a waveform.

\section{Related work}

A limited number of papers have provided a proof of concept that TTS without T is feasible \cite{DBLP:conf/icassp/MuthukumarB14,DBLP:conf/icassp/ScharenborgBBHM18}. We use this work as a baseline for the current challenge. This baseline uses out-of-the-box acoustic unit discovery \cite{DBLP:conf/sltu/OndelBC16} and an out-of-the-box speech synthesizer (Merlin, with the Ossian front end \cite{DBLP:conf/ssw/WuWK16}). Recent improvements on both the unit discovery and the synthesis side of the problem promise to improve on this baseline.

On the acoustic unit discovery side, several methods have been used (binarized autoencoders \cite{DBLP:conf/icassp/BadinoCFM14}, binarized siamese networks \cite{GLU2017}; a variety of speaker normalization techniques have been used to improve the categories \cite{heck2017feature}, among other techniques). On the speech synthesis side, waveform generation has recently seen great improvement (Wavenet \cite{DBLP:conf/ssw/OordDZSVGKSK16}, SampleRNN \cite{DBLP:journals/corr/MehriKGKJSCB16}, Tacotron 2 \cite{DBLP:conf/icassp/ShenPWSJYCZWRSA18}, DeepVoice3 \cite{DBLP:journals/corr/abs-1710-07654},  Transformer TTS \cite{DBLP:journals/corr/abs-1809-08895}, and others), some of these systems being open source.

Recent research shows that training ASR and TTS jointly with reconstruction losses can result in improvement in both  \cite{DBLP:conf/asru/TjandraS017}.
The ZR19 task is also similar to the voice conversion problem \cite{DBLP:journals/corr/abs-1711-11293}, in which the audio of a given source speaker is converted to a target speaker, without annotation (using VAEs \cite{DBLP:conf/apsipa/HsuHWTW16}, disentangling autoencoders \cite{DBLP:journals/corr/abs-1804-02812}, or GANs \cite{DBLP:conf/icassp/GaoSR18}). Recent work shows that such autoencoders can be trained with a discrete intermediate representation \cite{van2017neural,chorowski2019unsupervised}. The constraints ZR19 brings to the table are  (1) that the target synthesis voice be trained on a small, unannotated corpus; (2) that at test time, the system convert short recordings of novel source voices into this target voice; (3)  that the  embedding have good performance on a phoneme discriminability test; and  (4) that it have a low bit rate.

\section{Datasets}

\begin{table}

  \caption{Dataset statistics}
  \label{tab:datasets}
  \centering
  \begin{tabular}{llll}
    \toprule
    \textbf{Dataset} &\textbf{N speakers}      & \textbf{N utt.}    & \textbf{Duration} \\
        \midrule
Dev: Train Voice     & 1 M	    & 970	&   2h \\  
                & 1 F       & 2563  &	2h40\\
Dev: Train Unit Disc. &100	    & 5941  &	15h40\\
Dev: Train Parallel  &10 + 1 M   &	92  &	4.3min\\
                &10 + 1 F   &	98  &	4.5min\\
Dev: Test 	        &24	        &   455 &   28min\\ 
    \midrule
Sur: Train Voice     &	1 F       &	1862  &	1h30\\
Sur: Train Unit Disc. &	112       &	15340 & 15h\\
Sur: Train Parallel  &	15 + 1 F  &	150   &	8min\\
Sur: Test            &	15        &	405   & 29min \\
    \bottomrule
  \end{tabular}
\end{table}

We provide data for two languages. The Development language is English. Participants are to treat English as if it were low-resource, and not use existing resources. The Surprise language is Indonesian, for which the corpora are derived from those developed in \cite{sakti2008developmentsynth,sakti2008developmentasr}. Participants up to now have only been told that the Surprise language is an Austronesian language; all participants (past and future) are asked not to use resources from related languages. Only the Development language is to be used for  model development. Training on the Surprise language must  be exactly the same procedure applied to the Development language: hyperparameter optimization must be done on Development data or integrated into training. The goal is to build a system that generalizes out of the box to new languages.  

Four datasets are provided for each language (see Table \ref{tab:datasets}). The \textit{Train Voice} dataset contains either two talkers (Development) or one  (Surprise), and is for building an acoustic model of the target voice for speech synthesis. The \textit{Train Unit Discovery} dataset contains read text from multiple speakers, with around 10 minutes of speech from each speaker. These are for the discovery of speaker-independent acoustic units. The optional \textit{Train Parallel} dataset contains parallel spoken utterances from the Target Voice and from other speakers, intended to fine tune the task of voice conversion.\footnote{No submitted systems used this extra training set. In principle, we would rank such systems separately.} The \textit{Test} Dataset contains new utterances by new speakers. 
Requiring resynthesis in another voice allows us to exclude trivial solutions in which the original audio is returned unchanged.


\section{Metrics}
As shown in Figure \ref{fig:diagram-challenge}, participants feed each audio test item to their system and give two outputs for evaluation: at the end of the pipeline, participants submit the resynthesized audio file; in the middle of the pipeline, participants give the ``pseudo-text'' embedding used at the entry point to the synthesis component.\footnote{Additionally, participants could submit two auxiliary embeddings from earlier or later steps in the system’s pipeline in order to analyze the quality of non-binarized representations computed in the pipeline.}

The general form for the embedding is a sequence of vectors, each one of which can be seen as a ``symbol.'' The low bitrate constraint (see below)  favours a small, finite set of values for these vectors, as is the case for phoneme units in speech. The vectors might be one-hot  (each ``symbol'' coded as one on its own dimension, zero elsewhere), but are not limited to such representations, and can also be continuous-valued, to take advantage of the similarity structure of the embedding space. To reduce the bitrate, participants are nevertheless advised to quantize to a discrete subset of values.  The number of vectors for a given test file is not fixed, permitting participants to use a fixed frame rate, or to instead use ``character''-like encodings, in which successive identical symbols are collapsed and no notion of alignment is retained. The submission format does not distinguish between these cases. The machine evaluation calculates the bitrate and the embedding quality; human speakers of English and Indonesian are presented with the test sentences in an online experiment in order to evaluate the synthesis quality.



\subsection{Synthesis intelligibility, quality and speaker similarity}

\textit{Intelligibility} was measured by asking participants to orthographically transcribe the synthesized sentence. Each transcription was compared with the gold transcription using the Levenshtein distance, yielding a Character Error Rate (\textit{CER}).  The overall \textit{naturalness} of the synthesis was assessed on a 1 to 5 scale, yielding a Mean Opinion Score (\textit{MOS}).\footnote{The question posed was: \emph{Rate how natural the audio is, between  1 and 5 (1=very unnatural, 3 = neutral, 5=very natural).}} \textit{Speaker similarity} was assessed using a 1 to 5 scale. Sentences were presented in pairs (target voice, system voice).\footnote{The question posed was: \emph{Rate the similarity between the reference
 voice and the system voice, between 1 
and 5 (1 = very different voices,
 3 = neither similar nor different voices,
 5 = very similar voices).} Ten additional trials were included, for each participant, in which the reference voice was not the target voice but the source voice.} A  training phase occurred before each task. Three ``catch'' trials were included in the transcription, consisting of easy sentences from the original corpus not included in the rest of the experimental list, allowing us to detect participants that  failed to do the task.

Each participant performed the evaluation tasks in the same order (Intelligibility, Naturalness, Similarity), the overall evaluation lasting about one hour.  To avoid re-evaluation of the same sentence by the same participant, the sentences (types) were split into two disjoint subsets: one third for the Intelligibility task (62 for English, 49 for Indonesian), and two third for the Naturalness task (129 for English, 100 for Indonesian). The complete set of sentences was used in the Similarity task. In the Intelligibility and Naturalness tasks, all the sentences were seen by all subjects; in the Similarity task, a pseudo random one-third of the whole sentences was selected for each participant. Each sentence token was evaluated at least once with each system (the submitted, topline and baseline systems, as well as the original recordings).\footnote{In the Intelligibility task, each system was evaluated at least 70 times for English, and 148 times for Indonesian, with each combination of sentence and system seen at least once. In the Naturalness task, each system was evaluated at least 180 times for English, and 274 for Indonesian, with each combination of sentence and system seen at least 36 times for English and at least 68 times for Indonesian. In the Similarity task, each system was evaluated at least 89 times in English, and 120 times in Indonesian, and all possible combinations of sentence and system were seen by at least one participant.} English judges were recruited through Mechanical Turk. Indonesian judges were recruited through universities and research institutes in Indonesia. All were paid the equivalent of 10 USD. Only data from participants with $<$0.80 CER on catch trials were retained (Development: 35/35; Surprise: 68/69).

\subsection{Embedding bitrate and quality}

For the bitrate computation, each vector is processed as a character string. A dictionary of the possible values is constructed over the embedding file for the submitted test set. 
We thus assume that the entire test set corresponds to a sequence of vectors $U$ of length $n$: $U=[s_1,...,s_n]$. The bit rate for $U$ is then $B(U)=n \sum_{i=1}^{n}{\frac{p(s_i)log_{2}p(s_i)}{D}}$, where $p(s_i)$ is the probability of symbol $s_i$. The numerator is $n$ times the entropy of the symbols, which gives the optimal number of bits needed to transmit the sequence of symbols $s_{1:n}$. To obtain a bitrate, we divide by $D$, the total duration of $U$ in seconds.\footnote{A fixed frame rate transcription may have a higher bitrate than a ``textual'' representation due to the repetition of symbols across frames. For instance, the bitrate of a 5 ms framewise gold phonetic transcription is around 450 bits/sec and that of a ``textual'' transcription around 60 bits/sec.}

Since it is unknown whether the discovered representations correspond to particular linguistic units (phone states, phonemes, features, syllables), we evaluate unit quality with a theory-neutral machine ABX score, as in previous Zero Resource challenges \cite{versteegh2016zero,dunbar2017zero}. The machine-ABX discriminability between `beg' and `bag' is defined as the probability that $A$ and $X$ are closer than $B$ and $X$, where $A$ and $X$ are tokens of `beg', and $B$ a token of `bag' (or vice versa), and $X$ is uttered by a different speaker than $A$ and $B$. The global ABX discriminability score aggregates over the entire set of minimal pairs such as `beg'—`bag' to be found in the test set. The choice of the appropriate distance measure is up to the researcher. In previous challenges, we used by default the average frame-wise cosine divergence of the representations of the tokens along a DTW-realigned path.  We provide in the the option of instead replacing the cosine divergence with the KL divergence, or of instead using a normalized Levenshtein edit distance over the two sequences. We give ABX scores as error rates (0\% for the gold transcription, 50\% being chance). Each of the items compared in the ABX task is a triphone ([izi]-[idi], and so on), extracted from the test corpus. Each triphone item is a short chunk of extracted audio, to be decoded by the systems.\footnote{This differs from previous challenges. In previous challenges, longer audio files were provided for decoding, from which the representations of triphones were extracted after the fact using time stamps. In the 2019 edition, triphones are pre-extracted to allow for systems without fixed frame rates. Note that the topline phone-level language model performs sub-optimally on these files, because they begin with unlikely sequences of phones.} 

The baseline synthesis system we provide requires textual annotations as input,  forcing any  unit discovery systems using it to convert  embeddings into one-hot  (unstructured) representations before using them for synthesis. The loss of information incurred may be unwanted. In order to help showcase the unit discovery systems' performance, we also allow participants to submit up to two auxiliary embeddings, which may be, for example, the outputs of the system prior to quantization. These embeddings are submitted to the ABX and bitrate evaluations, and are represented in light grey in the figures presented below.

\section{Toplines and Baselines}

A baseline system is provided, consisting of a pipeline with an acoustic unit discovery system based on DPGMM \cite{DBLP:conf/sltu/OndelBC16,ondel2018bayesian}, and a parametric speech synthesizer based on Merlin \cite{DBLP:conf/ssw/WuWK16}. As linguistic features, we use contextual information (leading and preceding phones, number of preceding and following phones in current sentence), but no features related to prosody (TOBI), phonetic categories (vowel, nasal, and so on) or part-of-speech (noun, verb, adjective, and so on).  A topline system is also provided, consisting of an ASR system trained using Kaldi \cite{povey2011kaldi} on the original transcriptions. The acoustic model is a tri-state triphone model with 15000 Gaussian mixtures. The language model is a trigram phone-level language model.\footnote{A word-level language model gives better performance, but we use a phone-level language model in the interest of giving a fair comparison with the subword unit discovery systems asked for in the challenge.} Output is piped to the TTS system, which is also trained on the gold labels. The baseline system is provided in a container.

\section{Results}

Nineteen systems were submitted, from ten groups (see Table \ref{tab:systems}), using a variety of approaches.  Only a few (LI, CH and TJ) used an end-to-end framework. About half used a fixed-rate frame-based encoding (resulting in higher bitrate, but finer temporal information), and half a character-based encoding (a sequence of units not proportional in length to the audio sample). Unit discovery methods were diverse (k-means, DPGMM, binarized autoencoders). Most systems used a vocoder, and only one (CH) directly generated audio. 
We present highlights of the results as of April 2019. To access the current leaderboard, including audio samples, see \mbox{\url{www.zerospeech.com/2019/results.html}}. In all figures, systems are coded by short names as in Table \ref{tab:systems}, followed by the leaderboard entry number. Auxiliary embeddings for each submission are followed by A or B.


\begin{table}[t]
  \caption{Characteristics of the submitted systems}
  \label{tab:systems}
  \centering
  \setlength\tabcolsep{2 pt}
  \begin{tabular}{lccc}
    \toprule
\textbf{System}  & \textbf{End-to-end} & \textbf{Frame-based}  &\textbf{Generation}   \\
    \midrule
    PA \cite{pandia}     & no & no & Ossian \\ 
    HO (Horizon Robotics) & no & yes & ? \\
    FE  \cite{feng}      & no & no & Ossian \\
    LI \cite{liu}        &yes &yes & Inversion \\
    EL \cite{kamper}     & no &yes & FFTnet  \\
    NA \cite{nayak}   & no & no & Ossian\\
    CH (Cho et al.)      &yes & yes & Direct \\
    RA (Rallabandi et al)& no & no & Ossian  \\
    YU \cite{boun19zs}   & no & no & Ossian\\
    TJ \cite{sakti}  &yes &yes & Inversion  \\
    \bottomrule
  \end{tabular}
\end{table}

In previous challenges, the baseline ABX reference for subword embedding quality was calculated on MFCCs, and the topline reference was calculated on supervised posteriorgrams (calculated using DTW with  KL-divergence). Here we also provide scores for our unsupervised baseline, and for the topline ASR phone decoding (calculated on Levenshtein distance).

\begin{figure}[t]
  \centering
  \includegraphics[width=\linewidth]{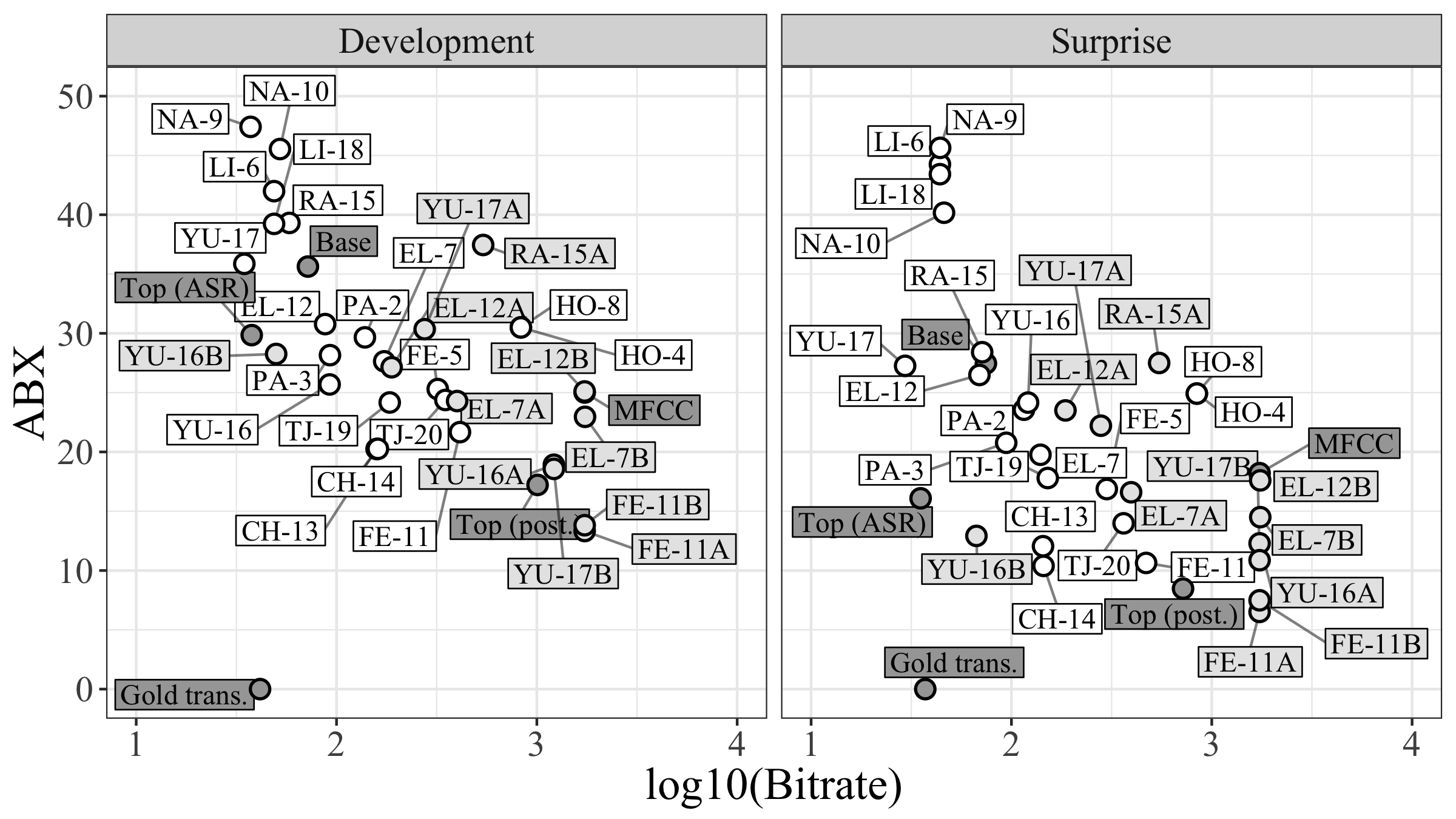}
  \caption{Embedding quality as a function of the log bitrate for the development language (\textbf{left}) and the surprise language (\textbf{right}). \textbf{Light grey boxes} represent auxiliary embeddings; \textbf{dark grey boxes} are our reference scores. Lower left is better.}
  \label{fig:abx}
\end{figure}

While systems in previous challenges generally improved over the MFCCs, the 2019 challenge was much more difficult, as seen in Figure \ref{fig:abx}. In the Surprise language, only five final embeddings achieved better performance than the MFCCs  (Dev: 25.01\%, Surprise: 18.21\%). The topline ASR posteriorgram ABX scores (Dev: 17.22\%, Surprise: 8.48\%) are comparable to previous challenges. The scores on the ASR decoding are worse than the posteriorgrams (Dev: 29.85\%, Surprise: 16.09\%)---and worse than the MFCCs, reflecting the fact that the supervised model was fairly simple.\footnote{The submitted non-discrete auxiliary embeddings showed generally better ABX scores than their discrete counterpart, at the expense of a higher bitrate, some of them even beating the supervised posteriors (FE: Dev: 13.82\%, Surprise: 6.52\%).}

Systems with high ABX error rates have low bitrates, while some systems with relatively high bitrates obtain better scores than the topline (the auxiliary embedding FE-11-A outperforms all others in the Surprise language; among final embeddings, CH-14 shows the best performance, between the decoding and posteriogram topline scores). This suggests that discretizing learned speech embeddings \emph{well} is hard. It also suggests that relatively dense representations, in spite of containing strictly more information than the phonemic transcriptions, are still useful in settings where only the linguistically relevant contrasts are necessary. The ``least discrete'' representations have a bitrate of the same order as the ASR posteriorgrams.

\begin{figure}[t]
  \centering
  \includegraphics[width=\linewidth]{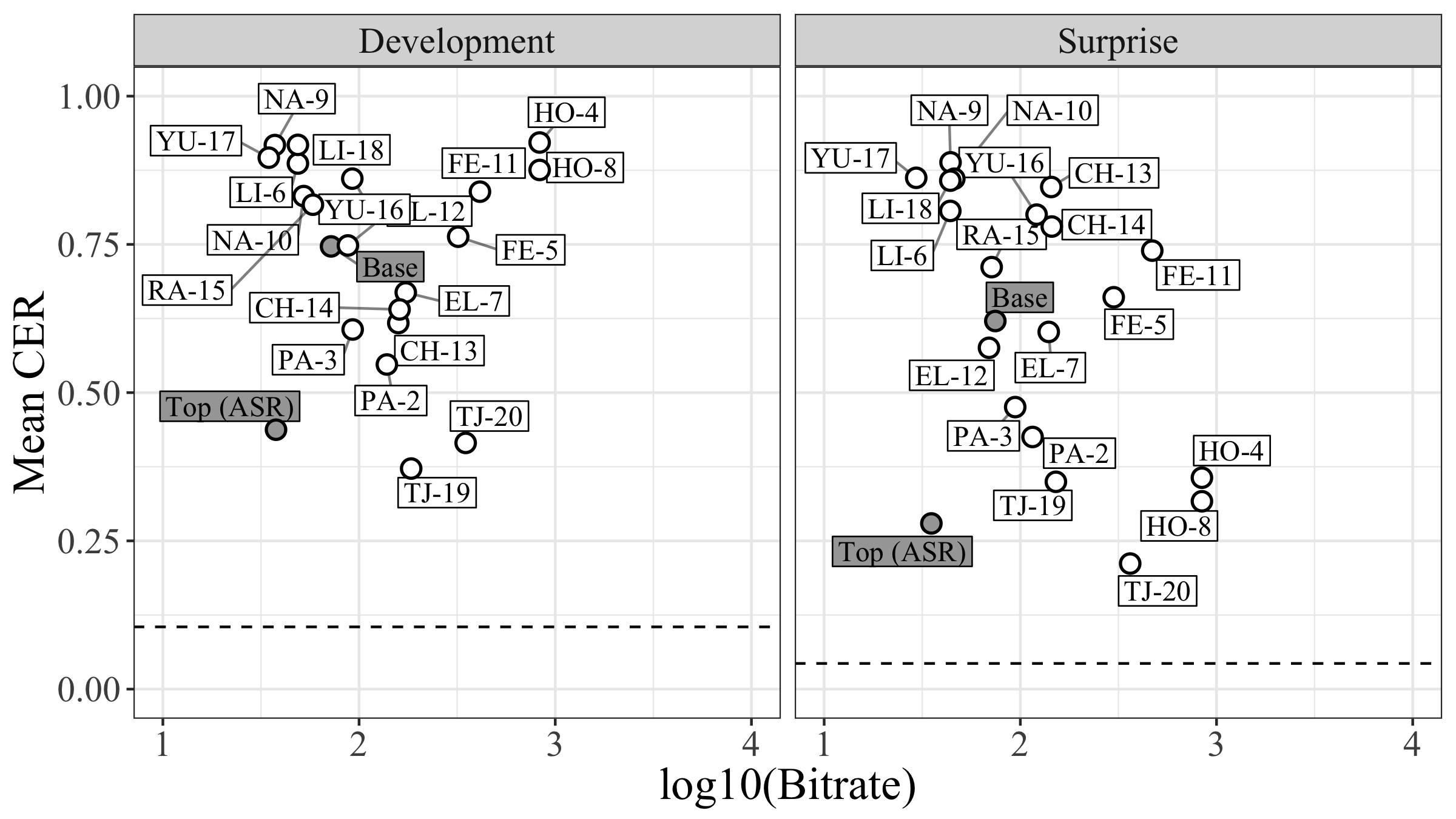}
  \caption{CER  as a function of the log bitrate for the development language (\textbf{left}) and the surprise language (\textbf{right}). \textbf{Dark grey boxes} are our reference scores. Lower left is better. \textbf{Dotted lines} are CER on original recordings.}
  \label{fig:cer}
\end{figure}

We take each system's mean over all trials, all participants and sentences pooled, for each of the three measures (CER, MOS, Similarity).\footnote{We also calculate bootstrap (N=10000) 95\% confidence intervals for each of these measures, for each system. We calculated half the width of the CI for each submission, in each of the two languages. We report the mean and the max (worst-case) CI half-width. For the Development language: CER, mean 0.04, max 0.07; MOS, mean 0.13, max 0.17; Similarity, mean 0.24, 0.28. For the Surprise language: CER, mean 0.03, max 0.05; MOS, mean 0.10, max 0.14; Similarity, mean 0.19, max 0.28. Differences between systems should be interpreted in light of these confidence intervals.}
We concentrate on the CER as a measure of synthesis quality (see below, and the online leaderboard, for information about the other measures). Figure \ref{fig:cer} shows that CER improves monotonically as a function of the bitrate.



MOS increases with lower CER (Dev: $r=-0.83$, Surprise: $-0.93$), as evaluators find intelligible speech more natural. Similarity increases with lower CER   (Dev: $r=-0.56$, Surprise: $-0.30$), but the correlation falls off substantially when the Gold audio is excluded (Dev: $r=-0.28$, Surprise: $-0.08$): some systems with good CER have low Similarity, while others that match the target voice well have high CER. The HO systems have particularly low Similarity, despite good CER. Low similarity and good CER may suggest trivial solutions, as discussed above.\footnote{Note, however, that Similarity only measures closeness to the target voice: low similarity does not distinguish between mere reproduction of the source voice, versus synthesis close to neither target nor source.} Similarity decreases as a function of bitrate (Dev: $r=-0.71$, Surprise: $-0.53$), suggesting that richer representations may code speaker information.

%

A central question motivating the 2019 edition is whether ABX phone discriminability  predicts success in a downstream task. There is indeed a monotonic relation between the ABX score and the CER ($r=0.54$ for Development, $0.44$ for Surprise). It is imperfect, as is  to be expected, since the quality of the synthesis module varies between systems.


\section{Discussion}

The Zero Resource Challenge 2019 shows that training  text-to-speech without textual annotations is possible, with quality on the level of a simple supervised comparison system. However, good performance is hard to achieve when the learned representations are ``text''-like. Systems with bitrates similar to the phonemic annotations tend to show poor performance: there is a  tradeoff between discretization and synthesis quality. We also find  the ABX phone discrimination measure on the discovered representations is  correlated with speech synthesis quality.

Not every setting requires ``textual'' representations. For synthesis in low-resource dialogue settings, high-bitrate representations may be sufficient. Still, the existence of the gold phonemic transcription---which gives reasonable synthesis even with our simple supervised baseline---shows there is substantial room for improvement with low-bitrate embeddings. One the reason for the worse performance of low-bitrate embeddings may be the use of the baseline Ossian synthesis, which requires textual input---a ``one-hot'' discretization that may lead to greater information loss than other methods of discretization---but this cannot be the whole explanation, as the gold transcriptions share this property. We suspect that future submissions to the 2019 Challenge, which remains open,  will explore these and other intriguing questions raised by the results.


%

\section{Acknowledgements}

ZR19  is supported by Facebook AI Research, a Microsoft Research grant,  INRIA, and  grants ANR-17-CE28-0009 (GEOMPHON), ANR-11-IDEX-0005 (USPC),  ANR-10-LABX-0083 (EFL), and ANR-17-EURE-0017. It is endorsed by SIG-UL, a joint ELRA-ISCA SIG on Under-resourced Languages. ZR19 is hosted on Codalab, an web-based platform for machine learning competitions.

\bibliographystyle{IEEEtran}

\bibliography{mybib}


\end{document}